\documentclass{article}

\PassOptionsToPackage{numbers, sort&compress}{natbib}
\usepackage[final]{neurips_2019}

\usepackage[utf8]{inputenc} 
\usepackage[T1]{fontenc}    
\usepackage{url}            
\usepackage{booktabs}       
\usepackage{amsfonts}       
\usepackage{nicefrac}       
\usepackage{microtype}      

\usepackage{footnote}
\usepackage{color,soul}
\usepackage{subcaption}
\usepackage[symbol]{footmisc}

\usepackage{graphicx}
\usepackage[most]{tcolorbox}
\usepackage[linesnumbered,ruled,vlined]{algorithm2e}
\usepackage{xcolor}
\usepackage{verbatim}
\usepackage{todonotes}
\usepackage{caption} 
\usepackage{enumitem}

\newcommand{\nonl}{\renewcommand{\nl}{\let\nl\oldnl}}

\newcommand\blfootnote[1]{%
  \begingroup
  \renewcommand\thefootnote{}\footnote{#1}%
  \addtocounter{footnote}{-1}%
  \endgroup
}

\title{Neural Architecture Evolution in Deep Reinforcement Learning for Continuous Control}

\author{
  \text{J\"org K.H. Franke$^{*,1}$}, \text{Gregor Koehler$^{*,2}$}, \text{Noor Awad$^1$}, \text{Frank Hutter$^{1,3}$} \\
  $^1$\text{University of Freiburg} \\ $^2$\text{German Cancer Research Center (DKFZ)} \\
  $^3$\text{Bosch Center for Artificial Intelligence}
}

\begin{document}

\maketitle

\maketitle

\begin{abstract}
  Current Deep Reinforcement Learning algorithms still heavily rely on handcrafted neural network architectures. We propose a novel approach to automatically find strong topologies for continuous control tasks while only adding a minor overhead in terms of interactions in the environment. To achieve this, we combine Neuroevolution techniques with off-policy training and propose a novel architecture mutation operator. Experiments on five continuous control benchmarks show that the proposed Actor-Critic Neuroevolution algorithm often outperforms the strong Actor-Critic baseline and is capable of automatically finding topologies in a sample-efficient manner which would otherwise have to be found by expensive architecture search.
\end{abstract}
\section{Introduction}

The\blfootnote{$^*$\text{Equal contribution. Correspondence to} \texttt{frankej@cs.uni-freiburg.de} \text{ and } \texttt{g.koehler@dkfz.de}.} field of Deep Reinforcement Learning (DRL) bears a lot of potential for meta-learning. DRL has recently achieved remarkable success in areas ranging from playing games \cite{Mnih13Atari, Silver16AlphaGo, alphastarblog}, locomotion control \cite{Lillicrap2015DDPG} and visual navigation \cite{Zhu16VisualNavi} to robotics \cite{Levine15Robotics, OpenAI18Dexterous}. 
However, all of these successes of DRL were based on manually-chosen neural architectures, rather than based on \emph{learned} architectures.

In this paper, we introduce a novel and efficient method for learning the architecture used in DRL algorithms for continuous control \emph{online}. To achieve this, we jointly learn the architecture of both Actor and Critic in a Q-Function Policy Gradient setting based on TD3 \cite{fujimotoTD3}. Specifically, our contributions are as follows:

\begin{itemize}[leftmargin=0.5cm, noitemsep, topsep=0pt]
    \item[--] We combine concepts from neuroevolution and off-policy RL training to evolve and train a population of actor and critic architectures in a sample-efficient way.
    \item[--] We propose a novel genetic operator based on network distillation~\cite{Hinton2015Distill} for stable architecture mutation.
    \item[--] Our method is the first to adapt the neural architecture of RL algorithms \emph{online}, based on off-policy training with the use of environment interactions from architecture evaluations shared in a global replay buffer.
    \item[--] Our method finds optimal architectures but only has a small overhead in terms of steps in the environment over a RL run with a single fixed architecture.
\end{itemize}

\section{Background}
Our proposed approach is based on Neuroevolution, a technique to optimize neural networks including their architecture using Genetic Algorithms (GAs) \cite{Stanley2002, Stanley09HyperNEAT, stanley2019designing}. First, a population of computation graphs with minimal complexity, represented by genomes, is created. Using genetic operators such as adding nodes or edges, additional structure is added incrementally. Different approaches in Neuroevolution usually differ in what is represented by the nodes and edges. In \cite{Stanley2002, Stanley09HyperNEAT}, each node represents an individual neuron and the edges determine the inputs to each neuron. Recent work extending Neuroevolution for larger network architectures used nodes to represent whole layers in a network, encoding the layer using its type-specific hyperparameters (e.g. kernel size in convolutional layers) \cite{miikkulainen2019evolving}. In this paper we follow a similar approach, encoding the network architecture with multi layer perceptrons  (MLPs). In contrast to the few existing works for learning neural architectures for RL (using blackbox evolutionary optimization~\cite{Chiang18AutoRL,Faust19EvolvingRew} or multi-fidelity Bayesian optimization~\cite{runge2019learning}), our approach optimizes the neural network architecture online, substantially improving sample-efficiency.

\section{Methods}
The foundation of Actor Critic Neuroevolution (ACN) is a genetic algorithm which evolves a population of $\mathcal{N}$ agents $\mathcal{P} = \{\mathcal{A}_1, \dots, \mathcal{A}_\mathcal{N} \}$. We associate each agent $\mathcal{A}_n = \lbrack f_n, \psi^a_n, \theta^a_n, \psi^c_n, \theta^c_n \rbrack$ with a fitness value $f_n$, along with topology descriptions $\psi^a_n$ and $\psi^c_n$, for actor and critic respectively, as well as their parameters $\theta^a_n$ and $\theta^c_n$.

For simplicity and comparability, we restrict the topology to standard MLPs for both actor and critic. After initializing the networks of each individual, we evaluate the actor MLP in the environment to obtain initial fitness values. With the initial fitness values in place, the evolution loop runs for $\mathcal{G}$ generations utilizing tournament selection \cite{Miller95tournament} for actor and critic individually to find the candidates for the next generation. Since mutation changes the actor, the critic can not be conditioned on the actor's behaviour and needs to be generally optimal. In the following sub-sections, we first explain the components of our algorithm and then how they are integrated in a single algorithm.

\subsection{Distilled Topology Mutation}
\label{sec:distill}
We introduce a novel mutation operator that acts on the topology of both the actor and the critic networks of the population. In order to mutate the actor and critic of an individual in a stable way, our proposed method operates in two stages. First, we jointly grow the topology of both networks in order to increase their capacity. With probability $p_L$, this growing mechanism appends another hidden layer of the same size as the previous last hidden layer to the respective networks; and with probability $1-p_L$, an existing hidden layer is chosen at random and a random number of new nodes are added to this layer. Both types of changes in topology are applied identically for actor and critic networks.

As the necessary initialization of the additional parameters introduced by growing the topology also changes both the critic’s estimate of the state-action values, as well as the policy of the actor, we propose a second stage of the topology mutation operator based on network distillation \cite{Hinton2015Distill, Rusu2016PolicyDistill}. Here we distill the parent's  behavior into the offspring, using data $\mathcal{D}^p = {\left(\boldsymbol{s_i}, \boldsymbol{q_i^p}\right)}_{i=1}^N$ consisting of $N$ states (or state-action pairs for critic distillation) sampled uniformly at random from the global replay memory, along with the parent network's outputs. This data is then used to perform gradient-based updates on the offspring network using the parent network's outputs as a target in a supervised learning setting:
\begin{equation}
    \mathcal{L}(\mathcal{D}^p, \boldsymbol{\theta^o)} = \sum_{i=1}^{|\mathcal{D}^p|} \left\| \boldsymbol{q_i^p} - \boldsymbol{\mu_{\theta}^o(\boldsymbol{s_i})} \right\|^2_2
\end{equation}
We apply a mean-squared-error loss (MSE) for both distillation updates on the offspring actor and critic, where $\boldsymbol{\mu_{\theta}^o}$ represents the respective offspring network with parameters $\boldsymbol{\theta^o}$. We use this additional step to stabilize the topology mutation operator, using the parent as a teacher to distill its knowledge into the offspring.

\subsection{Gradient-based Mutation}
\label{sec:safemutation}
We adopt a second mutation operator (SM-G-SUM) as one of two mutation operators used to evolve the actors in the population. This operator helps creating a more diverse set of actors in the population by altering the parameters of the actor network's layers. 
Neural network parameter mutations based on Gaussian noise can lead to strong deviations in behavior, often leading to deteriorated performance \cite{lehman2018safe, Bodnar19PDERL}. In order to stabilize the policies resulting from the mutation operator, we make use of the \textit{safe mutation} operator introduced in \cite{lehman2018safe}. This mutation approach scales the perturbations on a per-parameter basis, depending on the sensitivity of the network's outputs with respect to the individual parameter.

\subsection{Integration in Actor-Critic Neuroevolution}
To realize all benefits from the proposed genetic operators as well as the Actor-Critic training, we combine them in the Actor-Critic Neuroevolution (ACN) framework, see Algorithm \ref{alg:ACN} in the appendix. In the ACN framework, we integrate the two mutation operators described above in a standard GA loop, always mutating the selected candidates by either performing distilled topology mutation (with topology growth probability $p_G$) or gradient-based mutation (with probability $1-p_G$). After mutating, we add a network training phase for each individual following the setting of Twin-Delayed DDPG (TD3), performing multiple off-policy gradient updates making use of target policy smoothing and clipped double-Q learning \cite{Lillicrap2015DDPG, fujimotoTD3}. Due to the changes in the neural network architecture the training phase requires a re-initialization of the optimizer and a recreation of the target network at the start of each phase. 

The training of each individual can be performed in parallel, since each individual carries its own actor and critic. By adding this training phase, which uses the experiences from a global replay memory, each individual can benefit from a diverse set of policies exploring the environment during fitness evaluation. The training also improves the sample-efficiency of the GA as each offspring receives gradient-based updates, converging to high-reward solutions faster. This is in contrast to purely evolutionary approaches which have to explore the network parameter space in a highly inefficient manner.

\section{Experiments}
We evaluate ACN on 5 robot locomotion tasks from the Mujoco suite \cite{Todorov2012mujoco}. On these tasks, we compare the performance of a TD3 \cite{fujimotoTD3} baseline against two variations of ACN, one which evolves the architecture automatically and one with a fixed network architecture and parameter mutation only. This choice is motivated by the fact that TD3 is the algorithm we employ for each agent's individual training phase in ACN. To the best of our knowledge, this is the first work showing online architecture search on Mujoco tasks.
For all evaluated algorithms, we only use a single hyperparameter setting for all Mujoco environments to facilitate comparison with TD3, which was also evaluated with one fixed setting for all Mujoco environments. In terms of network architecture, the fixed architecture algorithms use two layers with 400 and 300 nodes respectively. In case of ACN evolving the topology, we start with a single layer of 64 hidden nodes, initialized using He initialization \cite{he2015delving}. 

\begin{figure}[!ht]
\centering
\begin{minipage}[b]{.32\textwidth}
  \centering
  \centerline{\includegraphics[width=1.0\linewidth]{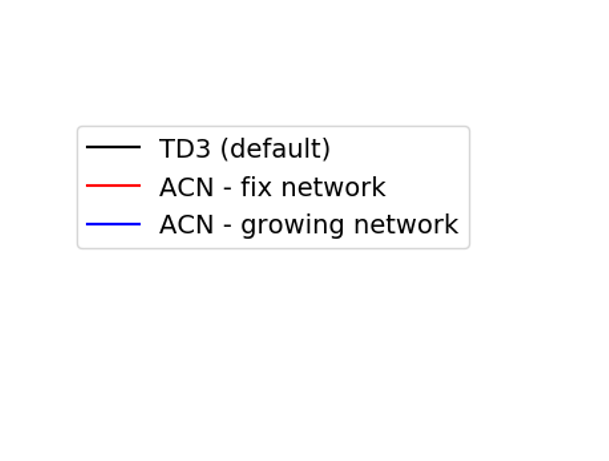}}
\end{minipage}
\hfill
\begin{minipage}[b]{.32\textwidth}
  \centering
  \centerline{\includegraphics[width=1.1\linewidth]{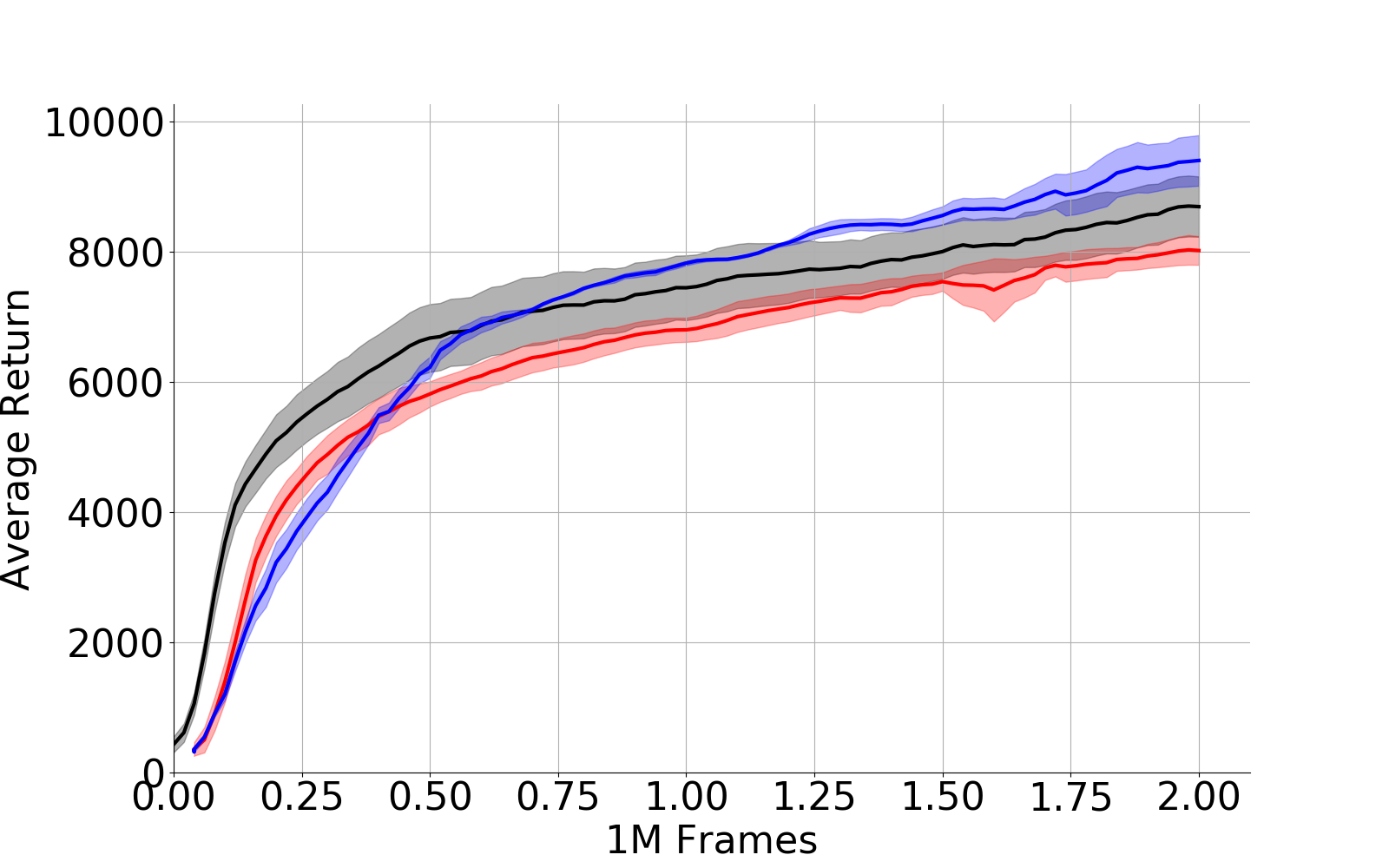}}
  \subcaption{HalfCheetah}\label{fig:comp_HalfCheetah}
\end{minipage}
\hfill
\begin{minipage}[b]{.32\textwidth}
  \centering
  \centerline{\includegraphics[width=1.1\linewidth]{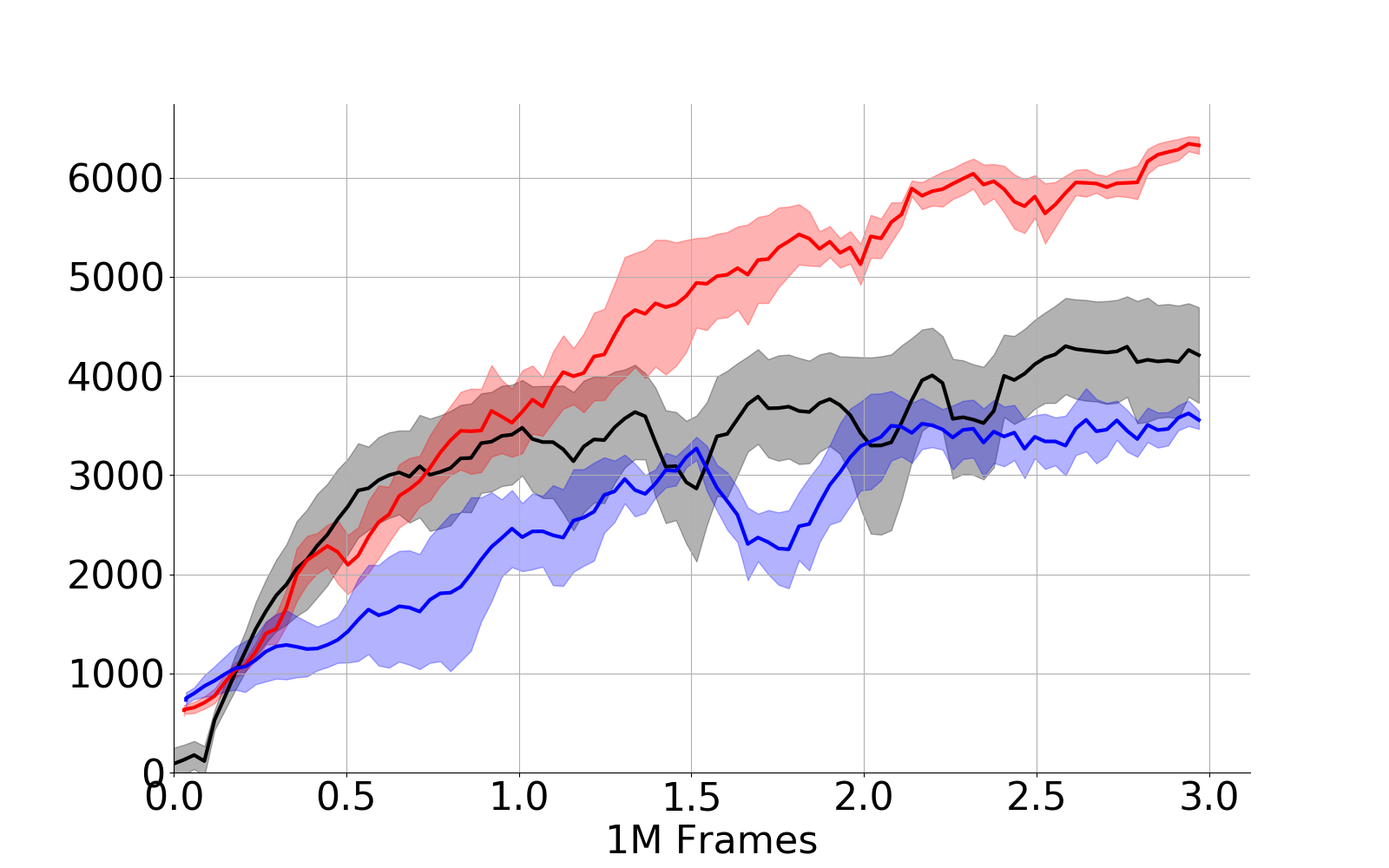}}
  \subcaption{Ant}\label{fig:comp_Ant}
\end{minipage}

\begin{minipage}[b]{.32\textwidth}
  \centering
  \vspace{0.2cm}
  \centerline{\includegraphics[width=1.1\linewidth]{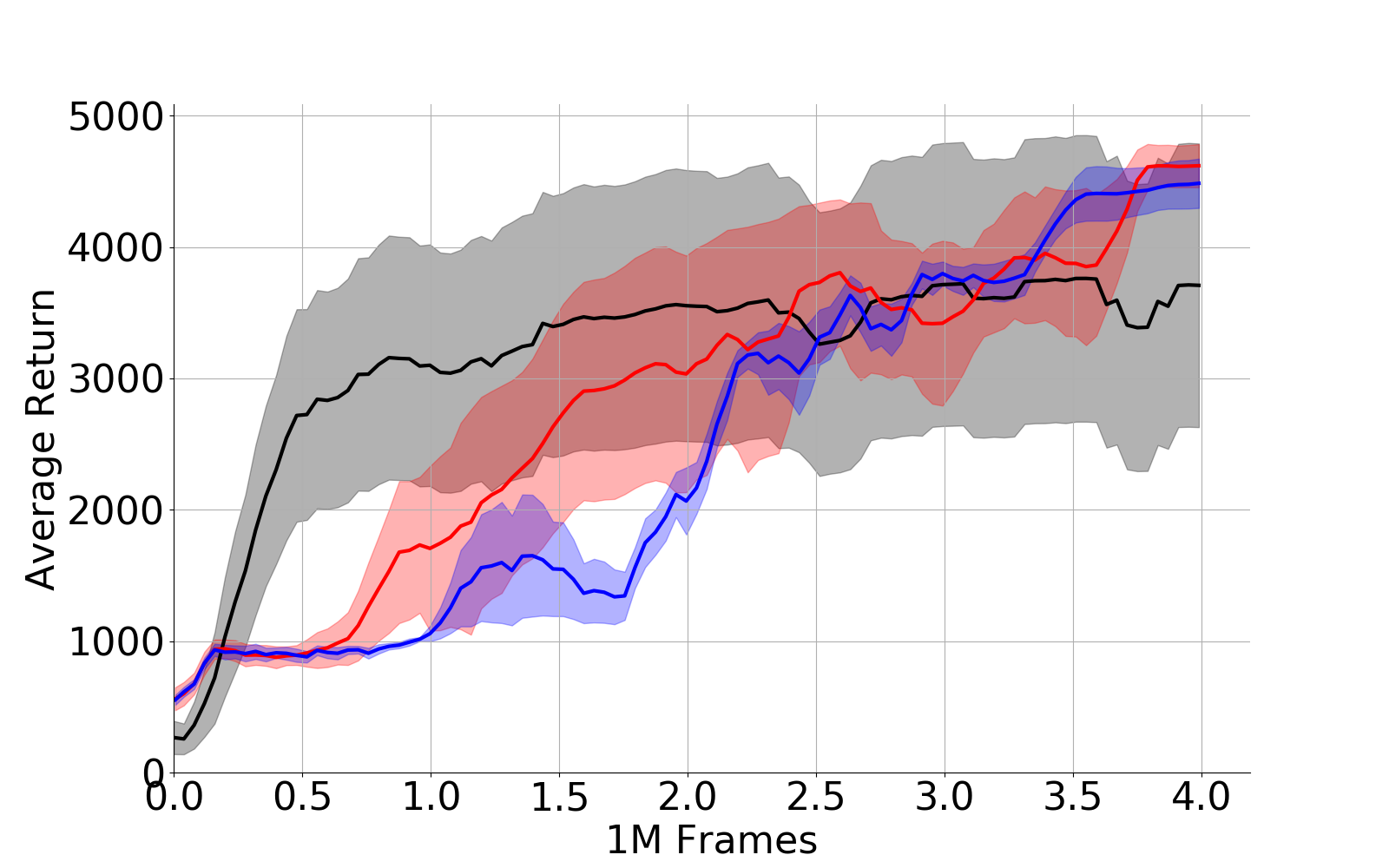}}
  \subcaption{Walker2d}\label{fig:comp_Walker2d}
\end{minipage}
\hfill
\begin{minipage}[b]{.32\textwidth}
  \centering
  \centerline{\includegraphics[width=1.1\linewidth]{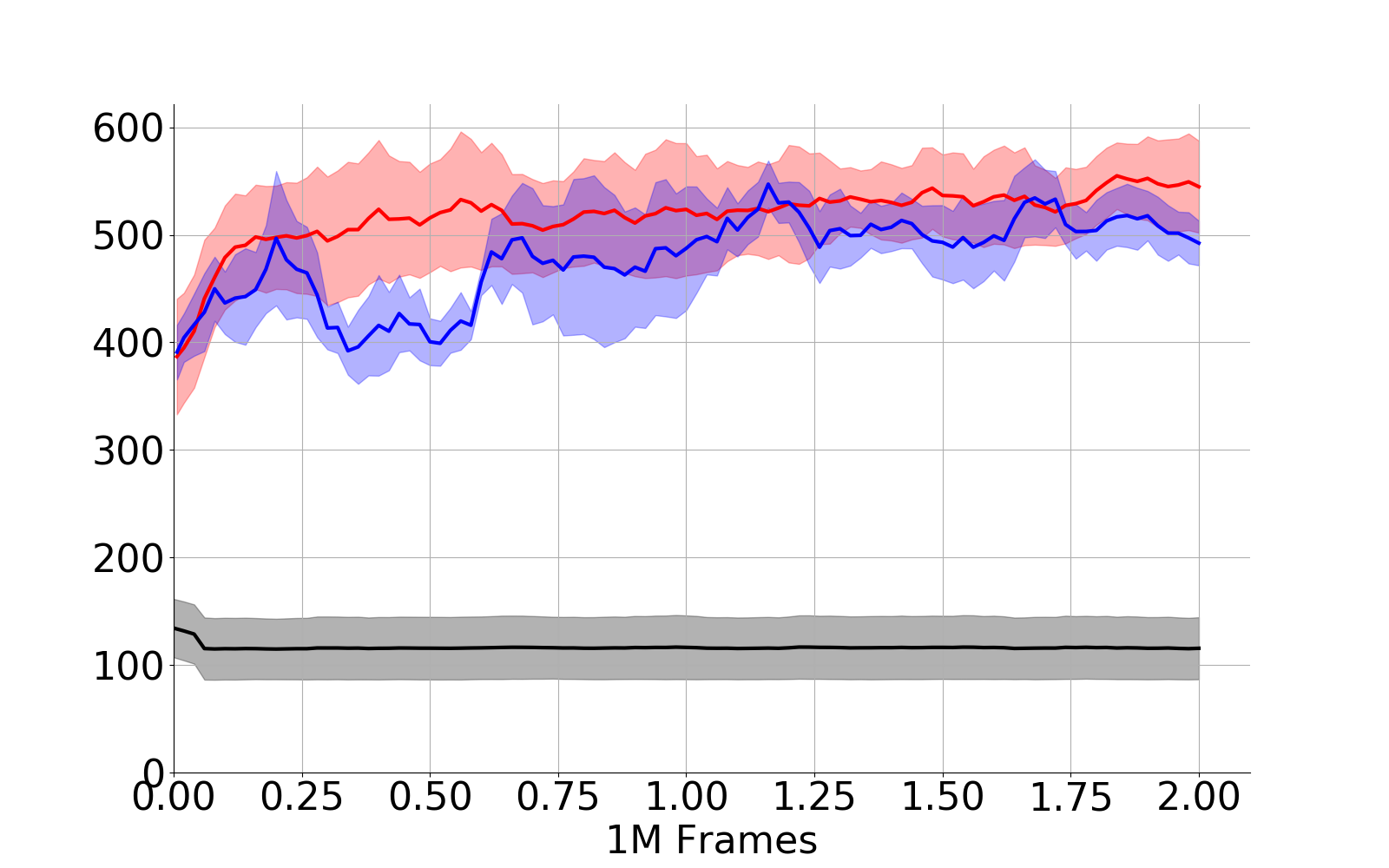}}
 \subcaption{Humanoid}\label{fig:comp_Humanoid}
\end{minipage}
\hfill
\begin{minipage}[b]{.32\textwidth}
  \centering
  \centerline{\includegraphics[width=1.1\linewidth]{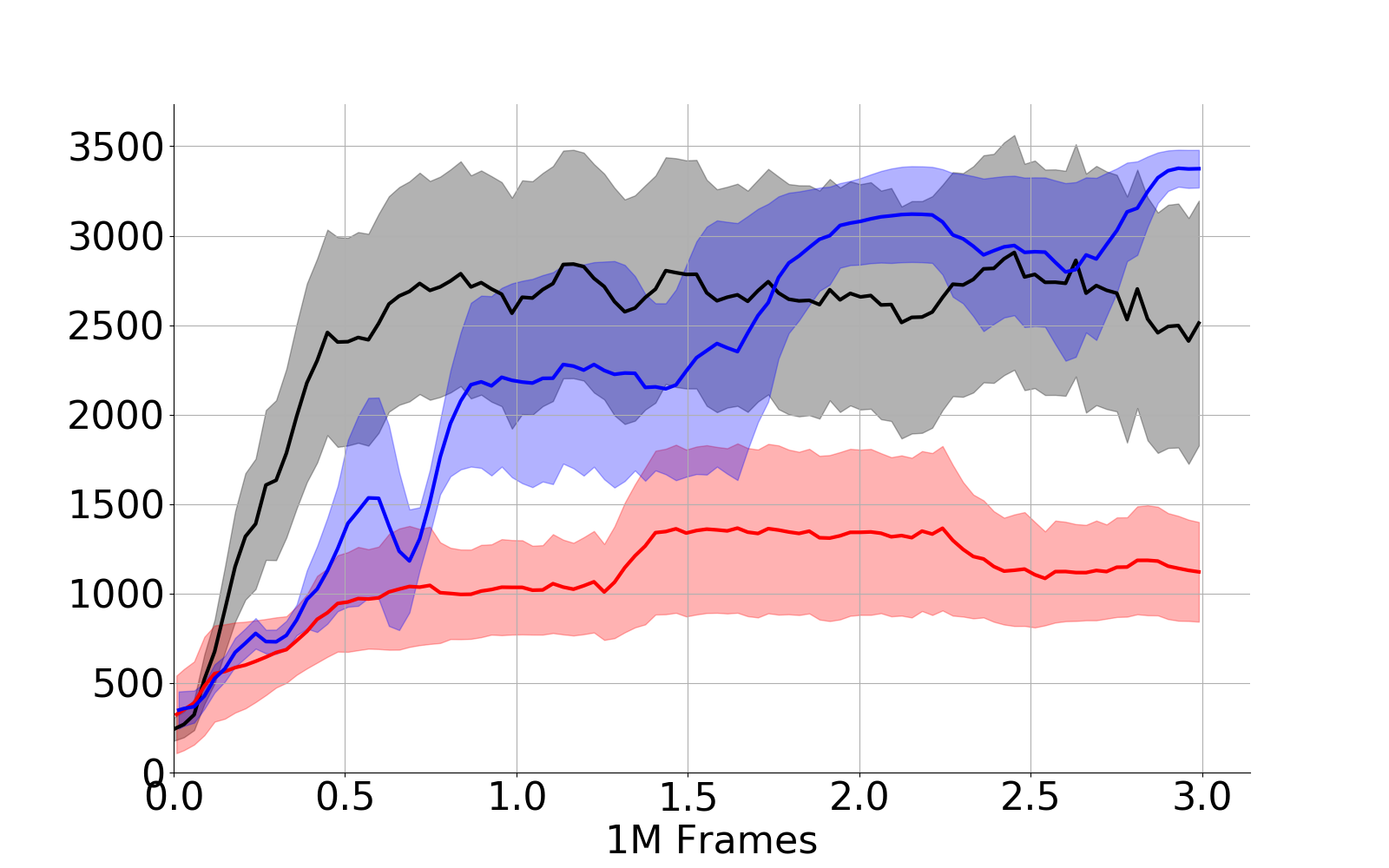}}
  \subcaption{Hopper}\label{fig:comp_Hopper}
\end{minipage}

\caption{Comparison of mean performance on continuous control benchmarks for ACN with fixed NN topology, with evolving neural network topology and TD3. We used two random seeds for ACN and five random seeds for TD3, the shaded area represents the standard error.}
\label{fig:ACN_vs_TD3_1}
\end{figure}

Figure \ref{fig:ACN_vs_TD3_1} shows that, at the end of the optimization, the two ACN variants perform on par with or better than TD3 on all evaluated continuous control tasks. The best architectures found by ACN and the architecture experiments with TD3 are given in Table \ref{table:topArch}.

Especially in the \textit{Humanoid} environment, the ACN algorithm shows a substantial improvement in performance, which can most likely be attributed to the exploratory nature of the algorithm, both in terms of NN topology and parameters. This is also reflected in the rather atypical architectures found by ACN for this task.
In \textit{HalfCheetah}, the final architectures found by ACN are smaller compared to the default architecture. This is consistent with the experiments in Appendix \ref{sec:topoExp} where a smaller size also outperforms the default architecture.
In \textit{Hopper}, ACN takes more environment steps to optimize the network architecture, but eventually catches up, again finding a smaller than usual network size consistent with the findings in Appendix \ref{sec:topoExp}.
The evolved network in \textit{Walker2d} also takes longer to optimize compared with a single TD3 run, but eventually outperforms TD3 with a smaller architecture.
The found architecture in \textit{Ant} only contains one layer and half the nodes compared to the TD3 default, but shows comparable performance. In this environment the fixed architecture variant of ACN outperforms TD3. This could be caused by the re-initialization and recreation of optimizer and target networks as shown in Appendix \ref{sec:reinitExp}.

\begin{table}[ht]
  \centering
  \begin{tabular}{l|ll}
    \toprule
    Environment & ACN & TD3 grid search\\
    \midrule
	Hopper & [136, 72]  & [200, 150]\\
	Ant & [276]  & [600, 450]\\
	HalfCheetah & [80, 80, 88]  & [600, 450]\\
	Humanoid & [672, 508]  & [400]\\
	Walker2d & [200, 144] & [600, 450]\\
    \bottomrule
  \end{tabular}
  \vspace{0.5em}
  \caption{Best actor architectures found by ACN compared with best performing TD3 runs.}
  \label{table:topArch}
\end{table}

The experiments show the capability of ACN to find suitable network architectures ranging from smaller architectures to larger ones, both in terms of number of layers and the individual layer sizes. ACN achieves this adding only a minor amount of computational cost.

Appendix \ref{sec:topoExp} shows experiments with different Actor/Critic NN architectures for TD3. These experiments show the significant impact network architecture choices can have on the algorithm's performance. We also evaluate the impact of re-initialization of the optimization algorithm and recreation of the target networks during training as it is applied during the ACN network training phase in Appendix \ref{sec:reinitExp}. For that experiment, we apply re-initialization of Adam and recreation of the target networks after each 10k training steps in TD3; it shows that the re-initialization and recreation does \emph{not} tend to have negative impact and sometimes even proves beneficial to the TD3 training.

\section{Conclusion}
This paper demonstrates how suitable neural network topologies of Actor and Critic networks can be found \textit{online}, while still showing performance comparable with state-of-the-art methods in robot locomotion tasks. We proposed the ACN algorithm, which combines the strengths of Neuroevolution methods with the sample-efficient training of off-policy Actor-Critic methods. To achieve this, we proposed a novel genetic operator which increases the network topology in a stable manner by distilling the parent network's knowledge into the offspring. Additionally, we augmented the GA with an off-policy Actor-Critic training phase, sharing collectively gathered environment interactions in a global replay memory.
Our experiments showed that ACN automatically finds suitable neural network architectures for all evaluated tasks which are consistent with strong architectures for these tasks, while only adding a small computational overhead over a single RL run with a fixed architecture.

Further work could investigate the impact of the mutation operator in RL training and why this combination of a GA and RL training often leads to a successful training of smaller topologies while achieving similar or even better performance compared to current RL algorithms.

\section*{Acknowledgments}
This work has partly been supported by the European Research Council (ERC) under the European Union’s Horizon 2020 research and innovation programme under grant no.\ 716721.

\setcitestyle{numbers}
\bibliographystyle{unsrtnat}
\bibliography{main}

\newpage
\appendix

\section{Neural Network Architecture Experiments for TD3}
\label{sec:topoExp}

\begin{figure}[!htb]
\centering
\vspace{-0.2cm}
\begin{minipage}[b]{.32\textwidth}
  \centering
  \centerline{\includegraphics[width=0.9\linewidth]{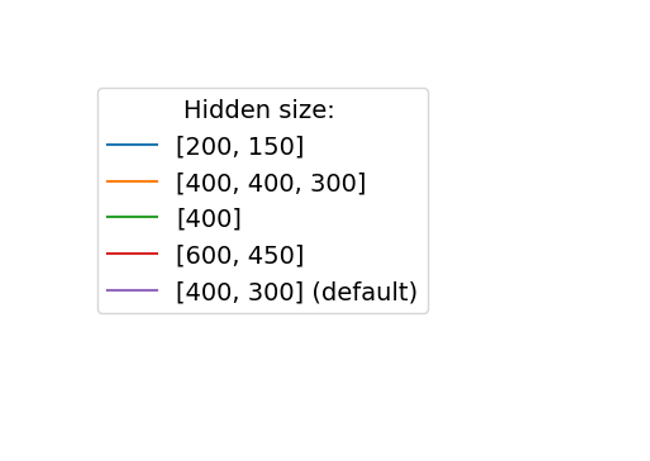}}
\end{minipage}
\hfill
\begin{minipage}[b]{.32\textwidth}
  \centering
  \centerline{\includegraphics[width=1.1\linewidth]{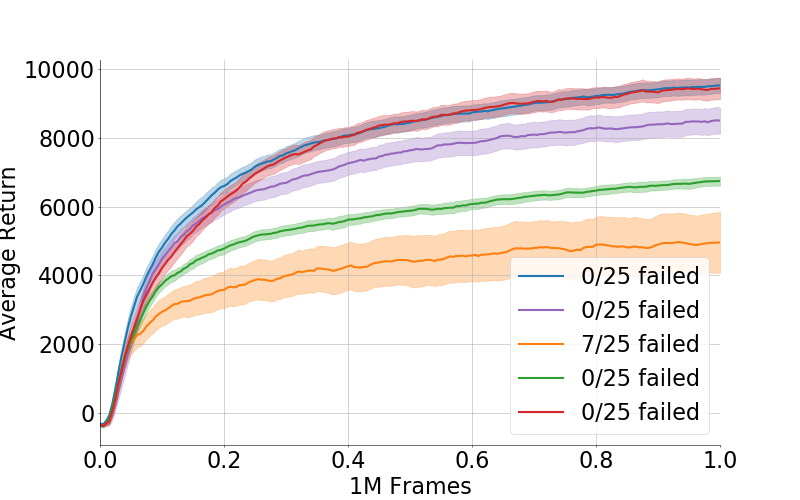}}
  \subcaption{HalfCheetah}\label{fig:comp_HalfCheetah2}
\end{minipage}
\hfill
\begin{minipage}[b]{.32\textwidth}
  \centering
  \centerline{\includegraphics[width=1.1\linewidth]{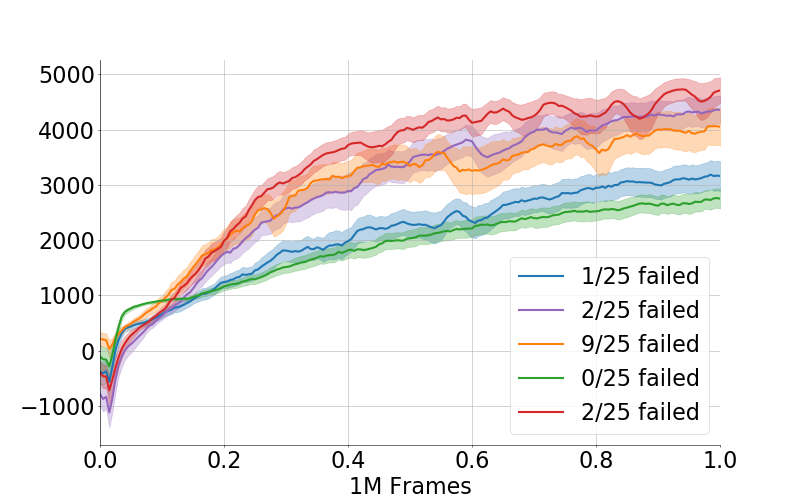}}
  \subcaption{Ant}\label{fig:comp_Ant2}
\end{minipage}
\begin{minipage}[b]{.32\textwidth}
  \centering
  \vspace{0.2cm}
  \centerline{\includegraphics[width=1.1\linewidth]{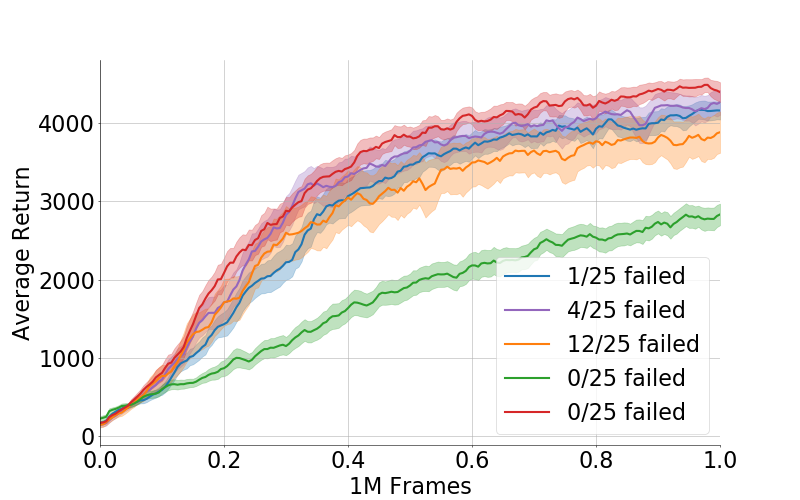}}
  \subcaption{Walker2d}\label{fig:comp_Walker2d2}
\end{minipage}
\hfill
\begin{minipage}[b]{.32\textwidth}
  \centering
  \centerline{\includegraphics[width=1.1\linewidth]{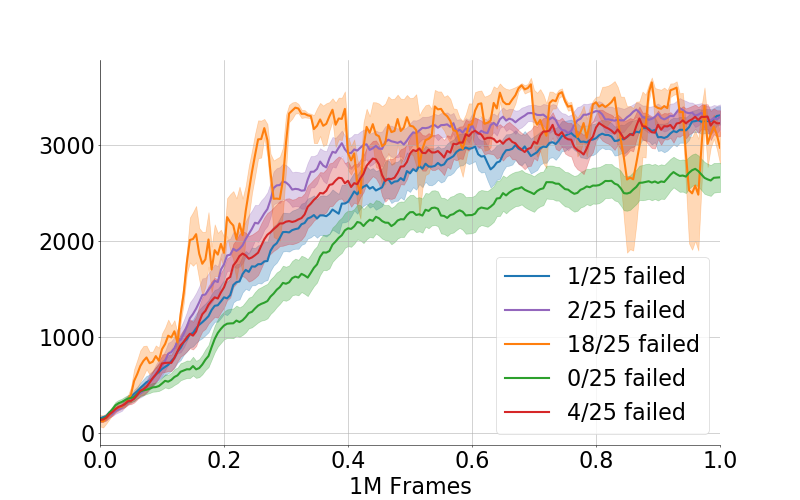}}
 \subcaption{Hopper}\label{fig:comp_Hopper2}
\end{minipage}
\hfill
\begin{minipage}[b]{.32\textwidth}
  \centering
  \centerline{\includegraphics[width=1.1\linewidth]{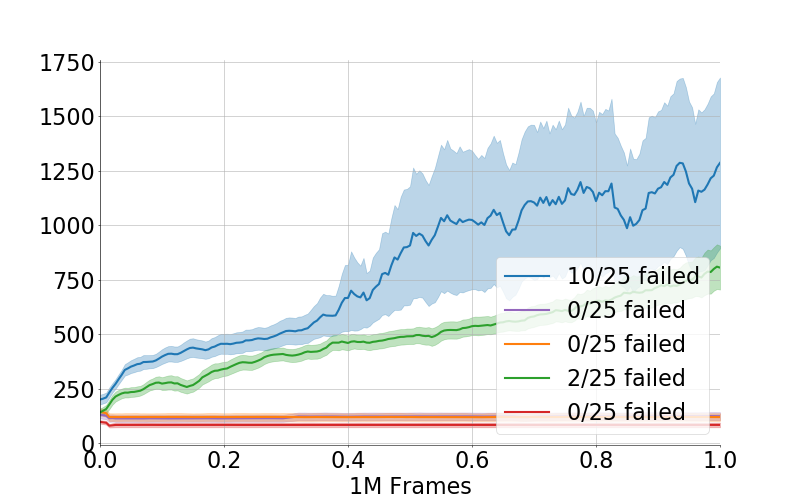}}
  \subcaption{Humanoid}\label{fig:comp_Humanoid2}
\end{minipage}

\caption{Comparison of the mean performance of TD3 with different neural network topologies on MuJoCo continuous control benchmarks. We used 25 random seeds, but we exclude and report the failed runs, where the average return was less than 5\% of the best run. The shaded area represents the standard error.}
\label{fig:TD3_topos}
\end{figure}

We evaluated different choices for the neural network architecture used for both actor and critic networks in the TD3 algorithm in figure \ref{fig:TD3_topos}, keeping all hyperparameters fixed as reported in the original work \cite{fujimotoTD3}. Figure \ref{fig:TD3_topos} shows the results for various neural network architecture choices. Perhaps unsurprisingly, the default architecture chosen in recent literature does not show the best performance on all environments. For example, in the \textit{Humanoid} environment, a simpler topology using only one hidden layer with 400 nodes performs substantially better, while in other environments like \textit{HalfCheetah, Ant} and \textit{Walker2d}, larger capacities like [600, 450] seem to be favorable. 

\section{ACN Algorithm}
\label{sec:ACNalgo}

A high-level description of the proposed ACN algorithm can be seen in Algorithm \ref{alg:ACN}, making use of tournament selection, Actor-Critic training based on TD3 and the combined mutation operator combining gradient-based mutation and the proposed distilled topology mutation (see Algorithm \ref{alg:mutation}).

\begin{algorithm}[!ht]
\caption{Actor-Critic Neuroevolution (ACN)
algorithm}
\label{alg:ACN}
    \SetAlgoLined
    \KwIn{population size $k$, number of generations $\mathcal{G}$}
    Initialize global replay memory $\Re$ \\
    Initialize $k$ individuals $\mathcal{A}_i := \{actor_i, critic_i, fitness_i=None \}$ as initial population $\mathcal{P}^*_0$ \\
    \For{$g=1$ \KwTo $\mathcal{G}$}{
        $\mathcal{P}_{g-1}$, $\mathcal{T}_e$ $\leftarrow$ \textsc{Evaluate}($\mathcal{P}^\#_{g-1}$) \\
        Store transitions $\mathcal{T}_e$ in $\Re$ \\
        $\mathcal{P}_{elite}$ $\leftarrow$ \textsc{TopK}($\mathcal{P}_{g-1}$) \\
        $\mathcal{P}_{selection}$ $\leftarrow$ \textsc{TournamentSelect}($\mathcal{P}_{g-1}$) \\
        $\mathcal{T}_s \leftarrow$ sample transitions from $\Re$ \\
        $\mathcal{P}^\#_{mutated} \leftarrow$ \textsc{Mutate}($\mathcal{P}_{selection}$, $\mathcal{T}_s$) \\
        $\mathcal{P}^\#_{trained}$ $\leftarrow$ \textsc{ActorCriticTraining}($\mathcal{P}^\#_{mutated},  \mathcal{T}_s$) \\
        $\mathcal{P}^\#_g \leftarrow$ $\mathcal{P}^\#_{trained} \cup \mathcal{P}_{elite}$ 
        }
\nonl\tcp{$ ^\#$Individuals are not evaluated in environment.}
\end{algorithm}

The \textsc{Evaluate} function performs a number of Monte-Carlo rollouts in the environment and determines the fitness value as the average cumulative reward across the performed rollouts. The selection operator \textsc{TournamentSelect} runs individual tournaments for all actors and critics of the new population, allowing for different Actor-Critic combinations in subsequent children. This choice is made to prevent actors from exploiting their critic's weakness over time, leading to undesired behavior.

After selection, the new population is mutated using transition batches from the global replay memory. With probability $p_{growth}$, the network architecture is grown. This is achieved either by appending a new layer of the same size as the last hidden layer of the actor (with probability $p_{add layer}$) or by choosing a number of additional nodes from the set given as a hyperparameter and adding this many nodes to any randomly chosen layer of the architecture. Both architecture growth operators perform identical architecture changes to both the actor and the critic, as indicated by \textsc{AddSameLayer} and \textsc{AddSameNodes} in Algorithm \ref{alg:mutation}. The set of possible additional node numbers can be found in Table \ref{table:hparams}. As this alteration of the network architectures changes the respective network's behavior, we perform network distillation (see section \ref{sec:distill}) updates using transition batches sampled uniformly at random from the global replay memory on both networks. With this additional step, we can distill the respective parent's knowledge into the offspring, thus enabling to grow the network architectures in a stable manner without requiring additional rollouts in the environment.\\
Alternatively to growing the network architectures, we mutate the individual actors of each agent in the population with probability $1-p_{growth}$, making use of the \textsc{SafeMutation} operator described in section \ref{sec:safemutation}. This mutation operator alters the individual's policy in a stable way, facilitating exploration in the environment.

\begin{algorithm}[!ht]
\caption{Mutation Algorithm}\label{alg:mutation}
    \SetAlgoLined
    \KwIn{population $\mathcal{P}$, transitions $\mathcal{T}$} 
    Initialize empty new population $\mathcal{P}_{mutated} = \{\}$ \\
    \For{each individual $i$ $\in$ $\mathcal{P}$ with actor $\mathcal{A}$ and critic $\mathcal{C}$}{
        \uIf{random number < $p_{growth}$}{
            \uIf{random number < new layer probability}{
            $\mathcal{A}_{grown}, \mathcal{C}_{grown} \leftarrow$ \textsc{AddLayer}($\mathcal{A}, \mathcal{C}$)\\
             }
            \Else{
            $\mathcal{A}_{grown}$, $\mathcal{C}_{grown}$ $\leftarrow$ \textsc{AddNodes}($\mathcal{A}, \mathcal{C}$)\\
            }
            $\mathcal{A}_{distilled} \leftarrow$ \textsc{DistillParent}($\mathcal{A}_{grown}, \mathcal{A}, \mathcal{T}$)\\
            $\mathcal{C}_{distilled} \leftarrow$ \textsc{DistillParent}($\mathcal{C}_{grown}, \mathcal{C}, \mathcal{T}$)\\
        }
        \Else{
            $\mathcal{A}_{mutated} \leftarrow$ \textsc{SafeMutation}($\mathcal{A}$,  $\mathcal{T}$) \\
        }
        $\mathcal{P}_{mutated} \leftarrow \textsc{Add} \{\mathcal{A}_{mutated}, \mathcal{C}_{distilled}\}$ 
        }
        \Return $\mathcal{P}_{mutated}$
\end{algorithm}

Each offspring created during the mutation phase is then trained individually using the \textsc{ActorCriticTraining} operator, which follows the off-policy gradient-based updates described in \cite{Lillicrap2015DDPG}, with the extensions introduced in \cite{fujimotoTD3}. The trained offspring, along with the elite determined as the best performing individual during evaluation, is then used as the next generation in the GA.

\section{Hyperparameters}
\label{sec:hparams}

All hyperparameters are kept constant across all environments. For the TD3 training, the same set of hyperparameters as reported in the original paper \cite{fujimotoTD3} were used. Table \ref{table:hparams} shows the hyperparameters used for ACN across all evaluated environments. All neural networks use the ReLU activation function for hidden layers and linear/tanh activations for critic and actor networks, respectively. We apply Layernorm \cite{Ba16Layernorm} after each hidden layer as it has proven beneficial for the stability and performance across all experiments in this paper.

\begin{table}[!ht]
  \centering
  \begin{tabular}{ll}
    \toprule
    Hyperparameter & Value\\
    \midrule
	Population size & 20\\
	Elite size & 5 \% \\
	Tournament size & 3 \\
	Network growth probability & 0.2 \\
	Add layer probability & 0.2 \\
	Add nodes probability & 0.8 \\
	Set of possible nodes added during layer growth & [4, 8, 16, 32] \\ 
	Network distillation updates & 500\\
	Network distillation batch size & 100 \\
	Network distillation learning rate & 0.1\\
	Safe mutation batch size & 1500\\
	Safe Parameter mutation standard deviation & 0.1 \\
    \bottomrule
  \end{tabular}
  \caption{Hyperparameters, constant across all environments.}
  \label{table:hparams}
\end{table}

\section{Experiments on Re-Initialization of Optimizers in TD3}
\label{sec:reinitExp}
\begin{figure}[!htb]
\centering
\begin{minipage}[b]{.32\textwidth}
  \centering
  \centerline{\includegraphics[width=0.9\linewidth]{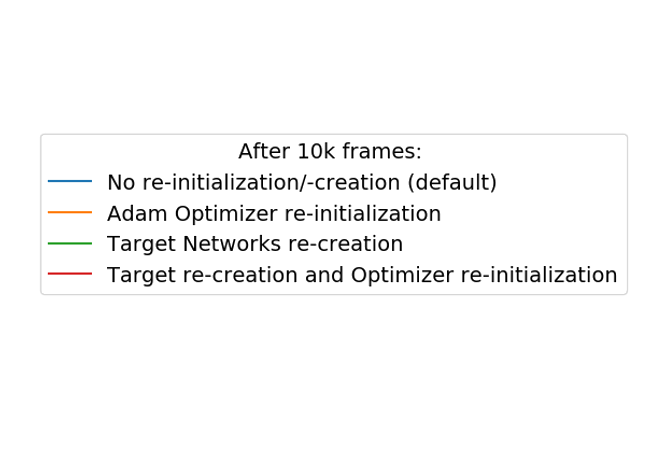}}
\end{minipage}
\hfill
\begin{minipage}[b]{.32\textwidth}
  \centering
  \centerline{\includegraphics[width=1.1\linewidth]{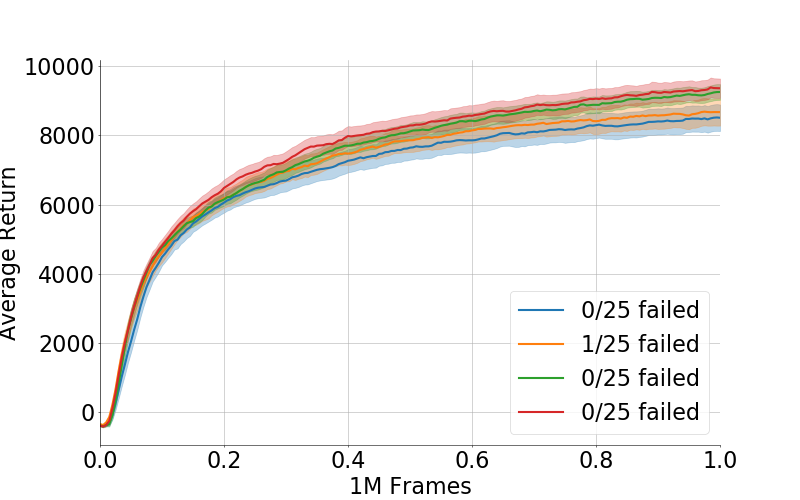}}
  \subcaption{HalfCheetah}\label{fig:comp_HalfCheetah3}
\end{minipage}
\hfill
\begin{minipage}[b]{.32\textwidth}
  \centering
  \centerline{\includegraphics[width=1.1\linewidth]{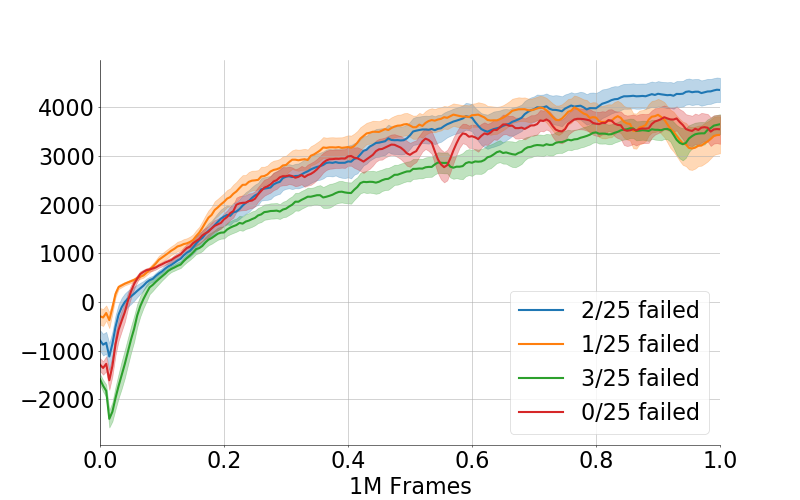}}
  \subcaption{Ant}\label{fig:comp_Ant3}
\end{minipage}
\begin{minipage}[b]{.32\textwidth}
  \centering
  \vspace{0.2cm}
  \centerline{\includegraphics[width=1.1\linewidth]{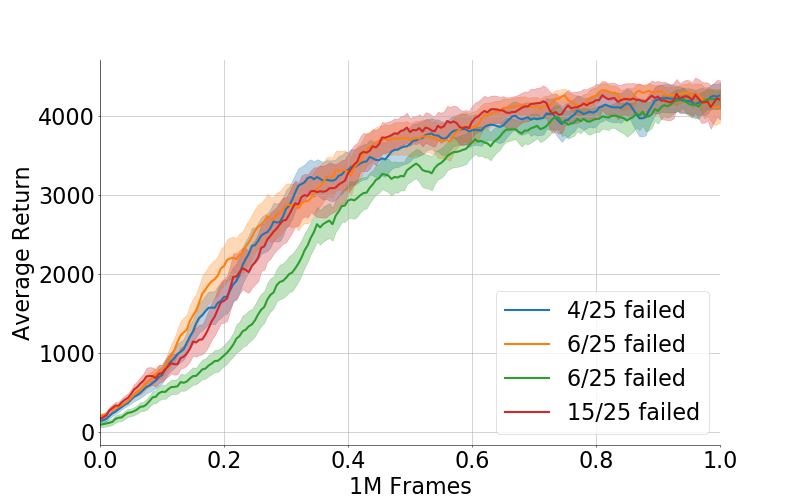}}
  \subcaption{Walker2d}\label{fig:comp_Walker2d3}
\end{minipage}
\hfill
\begin{minipage}[b]{.32\textwidth}
  \centering
  \centerline{\includegraphics[width=1.1\linewidth]{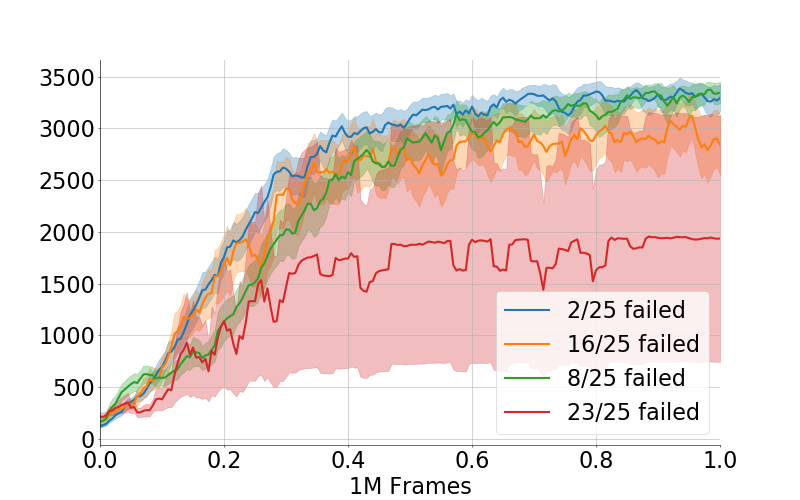}}
 \subcaption{Hopper}\label{fig:comp_Hopper3}
\end{minipage}
\hfill
\begin{minipage}[b]{.32\textwidth}
  \centering
  \centerline{\includegraphics[width=1.1\linewidth]{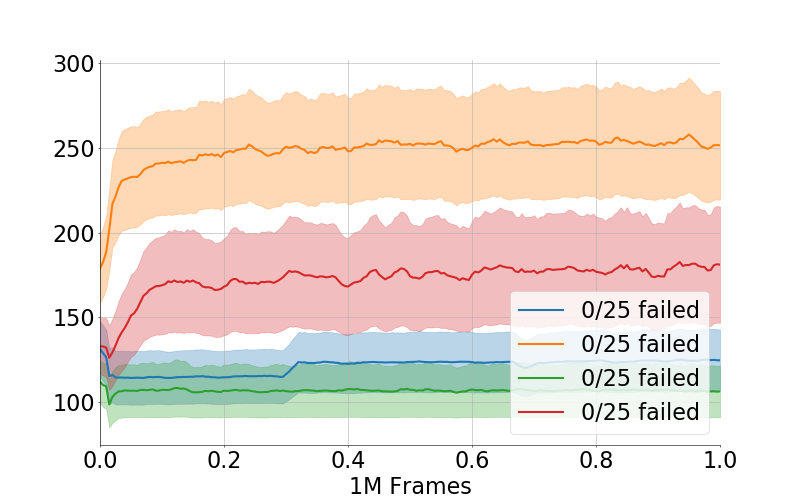}}
  \subcaption{Humanoid}\label{fig:comp_Humanoid3}
\end{minipage}

\caption{Comparison of the mean performance of TD3 with re-initializing the optimizer, re-create the target network or both after 10k frames on MuJoCo continuous control benchmarks. We used 25 random seeds, but we exclude and report the failed runs, where the average return was less than 5\% of the best run. The shaded area represents the standard error.}
\label{fig:TD3reinit}
\end{figure}

To assess the performance impact of both the re-initialization of optimizers as inevitably done in ACN, as well as using a new target network after a certain amount of steps, we evaluated different combinations in TD3. Figure \ref{fig:TD3reinit} shows the impact of different combinations on the performance of TD3 for the continuous control benchmarks used in this paper. Surprisingly, the default TD3 choice does not show the best performance in all environments, as might be expected. Rather, using the current state of the critic as the new target network from time to time seems to benefit performance.

\end{document}